\documentclass[wcp]{jmlr}

\usepackage{longtable}% for long tables
\usepackage{amsmath}
\usepackage{stfloats}

\usepackage{booktabs}
\makeatletter
\let\Ginclude@graphics\@org@Ginclude@graphics 
\makeatother

\jmlrvolume{}
\jmlryear{2022}
\title[Short Title]{MGADN: A Multi-task Graph Anomaly Detection Network for Multivariate Time Series}

 \author{\Name{Weixuan Xiong} \Email{weixuan@tju.edu.cn}\\
  \Name{Yun Qin} \Email{qinyunnn@163.com}\\
  \Name{Xiaochen Sun} \Email{xcsun@tju.edu.cn}\\
  \addr Tianjin University, Tianjin, China}

\begin{document}
\maketitle

\begin{abstract}

 Anomaly detection of time series, especially multivariate time series(time series with multiple sensors), has been focused on for several years. Though existing method has achieved great progress, there are several challenging problems to be solved. Firstly, existing method including neural network only concentrate on the relationship in terms of timestamp. To be exact, they only want to know how does the data in the past influence which in the future. However, one sensor sometimes intervenes in other sensor such as the speed of wind may cause decrease of temperature. Secondly,  there exist two categories of model for time series anomaly detection: prediction model and reconstruction model. Prediction model is adept at learning timely representation while short of capability when faced with sparse anomaly. Conversely, reconstruction model is opposite. Therefore, how can we efficiently get the relationship both in terms of both timestamp and sensors becomes our main topic. Our approach uses GAT, which is originated from graph neural network, to obtain connection between sensors. And LSTM is used to obtain relationships timely. Our approach is also designed to be double headed to calculate both prediction loss and reconstruction loss via VAE(Variational Auto-Encoder). In order to take advantage of two sorts of model, multi-task optimization algorithm is used in this model.
\end{abstract}

\section{Introduction}

A time series refers to a series of sequences arranged in the order of time occurrence. Time series are now widely used in speech analysis, noise removal, and stock market analysis. Multivariate time series is those time series having multiple channels. Through time series analysis, we can also find potential connections and numerical features implicit in these time series data, so that we can better understand various scientific theories, and we can also discover social phenomena from the data. Nature time series anomaly detection can help us locate anomalies in time, including systemic errors, disasters, and hardware and software damages. And use this to timely check and fill gaps to avoid the expansion of losses. Therefore, how to establish an accurate and efficient anomaly detection model is also an important topic in the field of time series analysis.

In real world, many systems product huge amount of inter-related multivariate time series data. For instance, in climate prediction system, different climate sensors receive different kinds of climate data such as temperature, moisture and wind speed etc. Apparently, those data collected by different sensors is related to each other which means one change can result in others' change. Most relationships between sensors are non-linear. For example, the surge of wind speed or sudden change of wind direction may bring massive cold air which in turns cause temperature decease. This pattern can't be learned only from previous temperature. 

In the past, traditional methods such as ARIMA and machine learning methods such as SVM(Support Vector Machines) are used to detect anomalies. Nevertheless, with the rapid growth of dimentionality and complexity of data from more sensors, traditional time series analysis methods are not appropriate anymore. Thus, deep learning methods become more and more popular and practical because of their strong ability of fitting different kinds of data. In terms of time series anomaly detection, two major deep learning methods are prediction model and reconstruction model. Fundamentally, AE(Auto Encoder) is a fairly practical model which use reconstruction error to discover the occurrence of anomaly. Its upgraded version-VAE(Variable Auto Encoder) regulate the distribution of latent space and use Kullback-Leibler divergence and binary cross entropy to calculate loss function which obviously strengthen its reconstruction power. More recently,  Generative Adversarial Networks (GANs), which is also acted as a reconstruction model, is used in the area. Construction models prove their competence to detect anomalies especially when anomalous labels are sparse, however proves their weakness when excavating timely relationships. On the contrary, prediction models such as RNN or LSTM-based approaches take full advantage of their sequential input pattern to learn timely relationships.
But when faced with sparse anomaly, the accuracy of prediction models decrease. In addition both two models cannot  explicitly learn the relationships between sensors. This limits their power to detect and explain deviations from such relationships when anomalous events happen.

In order to better learn the connections between sensors and make it contribute to the power of anomaly detection, Graph Neural Networks(GNNs) \cite{defferrard2016convolutional} should be considered. GNNs, including graph convolution networks (GCNs) \cite{kipf2016semi}, graph attention networks (GATs) \cite{velivckovic2017graph}, have shown its great ability to learn relationships between in graph structured data. However, most GNNs take data with pre-defined graph as input which is not applicable in multivariate time series data. Therefore, how to utilize GNNs in dealing with multi-time series without any prior information becomes our major topic. The first approach that fits GNNs on time series analysis is MTGNN proposed by \cite{wu2020connecting}.While MTGNN is for time series prediction, several GNNs approaches came up for not only prediction, but for anomaly detection or classification. However, they are either prediction model or reconstruction model, in other words, one headed model.

Thus, in this paper, we proposed our MGADN: A Multi-task Graph Anomaly Detection Network, which combines both prediction model and reconstruction model and uses GNNs to discover inherent relationship between sensors. MGADN have four main components:\textbf{(1).Shared Weight Layer:} provide shared parameters of the whole model and acquire timely pattern through certain designed structure.\textbf{(2).Construction head:} using VAE to reconstruct input and calculating reconstruction error. \textbf{(3).Prediction head:} using GNNs to learn graph from multi-time series and slide window methods to make prediction and calculating prediction error(deviation between ground truth and predicted value).\textbf{(4).Multi-task Optimization and Deviation Scoring:} calculate loss from two heads with multi-task optimization methods and identifies and explains deviations from the learned sensor relationships in the graph.

In conclusion, this work makes contributions below:
\begin{itemize}
\item Compared with the existing one-head model such as MTGNN \cite{wu2020connecting} and GDN \cite{deng2021graph}, we propose MGADN, a new two-headed and graph neural network based approach that exploit advantage of both prediction and reconstruction model and learn inter-relationship between sensors. The GNNs module can automatically get graph structure from dataset itself.

\item MGADN is modeled as a multi-task optimization method. When dealing with prediction loss and reconstruction loss, this multi-task learning assignment is transferred into multi-task optimization problem by calculating Pareto Optimality. This is for solving conflict between prediction task and reconstruction task to a certain extent.

\item We conduct experiments on three open datasets with ground truth anomalies. Our results demonstrate that MGADN detects anomalies more accurately than baseline approaches. Some other experiments such as ablation study and visualization work will prove efficiency of every individual module.

\end{itemize}

%\subsection{Subsection Title}
% A figure in Fig.~\ref{fig:spiral}. Please use high quality graphics for your camera-ready submission -- if you can use a vector graphics format such as \texttt{.eps} or \texttt{.pdf}.
% \begin{figure}[htp]
% \begin{center}
% \includegraphics[width=1\textwidth]{MGADN.pdf}
% \caption{A spiral.}\label{fig:spiral}
% \end{center}
% \end{figure}

\section{Related Work}
In this section, we will introduce several model related to multivariate time series modelling. In addition, since our approach is combined by graph neural network and multi-task learning, methods about these two topics will also be summarized in this section.

\textbf{Multivariate Time Series Modelling:}  The earliest time series analysis method is the auto-regressive (AR) model proposed by British statistician \cite{udny1927method}. In 1931, mathematician \cite{walker1931periodicity} established the Moving Average Model (Moving Average Model, MA). In the 1970s, the famous American statistician \cite{box2015time}, scientists who made outstanding contributions to time series analysis, proposed a method for abnormal detection of non-stationary time series. The non-stationary time series is transformed into a stationary series through appropriate order operations, and then the method of ARMA model is used. That is, the differential autoregressive moving average model (Autoregressive Integrated Moving Average Model, ARIMA). In 1980, \cite{lutkepohl2013vector} proposed the vector autoregressive model (Vector Autoregressive Models VAR). The VAR model is a generalization of the AR model. Since the AR model can only be applied to the scenario of univariate time series analysis, it has great limitations, and VAR extends the AR model to multivariate time series analysis.

Since 20s, the rapid rise of deep learning has brought more attention to neural network methods. \cite{zaremba2014recurrent} proposed Neural Network (Recurrent Neural Network RNN), as a cyclic feedback neural network framework, is often used in the processing of sequence data. It can fully capture the correlation in time series across time stamps. And its revised versions such as  Long Short Term Memory Networks(LSTMs) and Gated Recurrent Unit Networks(GRUs) are proved to be practical in multivariate time series analysis \cite{karim2017lstm, kumar2018long}. Generative Adversarial Networks(GANs) \cite{li2019mad, zhou2019beatgan}, Variable Auto-Encoder(VAE) and other reconstruction model are majorly used in anomaly detection.

\textbf{Graph Neural Networks:} In recent years, GNNs are considered as good methods to deal with graph structured data. The iterative mode of GNNs \cite{defferrard2016convolutional} is that the hidden state of one node is influenced by its neighbors and its own previous hidden state. Graph Convolution Networks(GCN) \cite{kipf2016semi} defines Fourier transformation on graph and upgrade its parameters by absorbing nodes' one-step neighbors' representation. Graph Attention Networks(GAT) \cite{velivckovic2017graph} include the thesis of attention which calculate attention weight of each neighbors to get the importance rank of them during aggregation process. To move forward a single step, some works prove the ability of GNNs on modelling large-scale multivariate time series data. For instance, STGCN proposed in 2017 have succeeded in predicting traffic flows \cite{yu2017spatio}.

With regard to anomaly detection, GDN, which is a prediction model, proposed by \cite{deng2021graph} use GATs to capture relationship between sensors. Moreover, GNNs can also be fixed into reconstruction model like VAE or GAN.
\cite{li2021stacking} proposed Stackvae-G which combines VAE and GNNs to detect anomaly in multi-time series.

\textbf{Multi-task Learning:} First we give the definition of Multi-task Learning(MTL) from \cite{zhang2018overview}:

\begin{itemize}
    \item \textbf{Definition 1 MTL:} 
    Given $m$ learning tasks $\left\{\mathcal{T}_{i}\right\}_{i=1}^{m}$ where all the tasks or a subset of them $i=1$ are related but not identical, multi-task learning aims to help improve the learning of a model for $T_i$ by using the knowledge contained in the $m$ tasks.
\end{itemize}

MTL can be divided into two categories: \textbf{hard parameter sharing} and \textbf{soft parameter sharing} \cite{ruder2017overview}. Hard parameter sharing method set hidden layers which have shared parameters between all tasks. In soft parameter sharing on the other hand, each task has its own model with its own parameters. MGADN will use hard parameter sharing method.
On the other hand, to balance different tasks, there are different ways of training set. One may set model train different tasks together with the a loss that is the weighted sum of different losses from diversed tasks. Another way is to train different tasks one by one which means several times of loss backward. \cite{sener2018multi} treated MTL as a multi-task optimization problem, and proved their methods practical.

\section{Proposed Model}

\subsection{Problem Statement}
In this work, our input of model is a multivariate time series dataset with $N$ sensors and $T$ time steps. We can denote input as $\left [  h^{1},h^{2},..., h^{T}\right ]$. With certain time stamp $t$, $h^{t}\in R^{N}$. With respect to prediction model, our task is to use slide window with its size $d$ to predict future value. The goal projection is: $f:\left [  h^{t+1},h^{t+2},..., h^{t+d}\right ]\longrightarrow \hat{h}^{t+d+1}$. As for reconstruction model, we can also use slide window with the same size $d$. The reconstruction goal will be $g:\left [  h^{t+1},h^{t+2},..., h^{t+d}\right ]\longrightarrow \left [  \hat{h}^{t+1},\hat{h}^{t+2},..., \hat{h}^{t+d}\right ]$. Moreover, we will calculate deviation score through prediction and reconstruction. Specifically, the deviation between prediction or reconstruction and ground truth will be transferred into error score which act as the criterion for anomaly diagnoses. That is, with a certain time $t$, the output of the model will be binary value $score(t)$, i.e, $score(t) \in \{0,1\}$, where $score(t)=1$ means occurrence of anomaly on time $t$.

\subsection{Overview of MGADN}
\begin{figure}[htp]
\begin{center}
\includegraphics[width=1\textwidth]{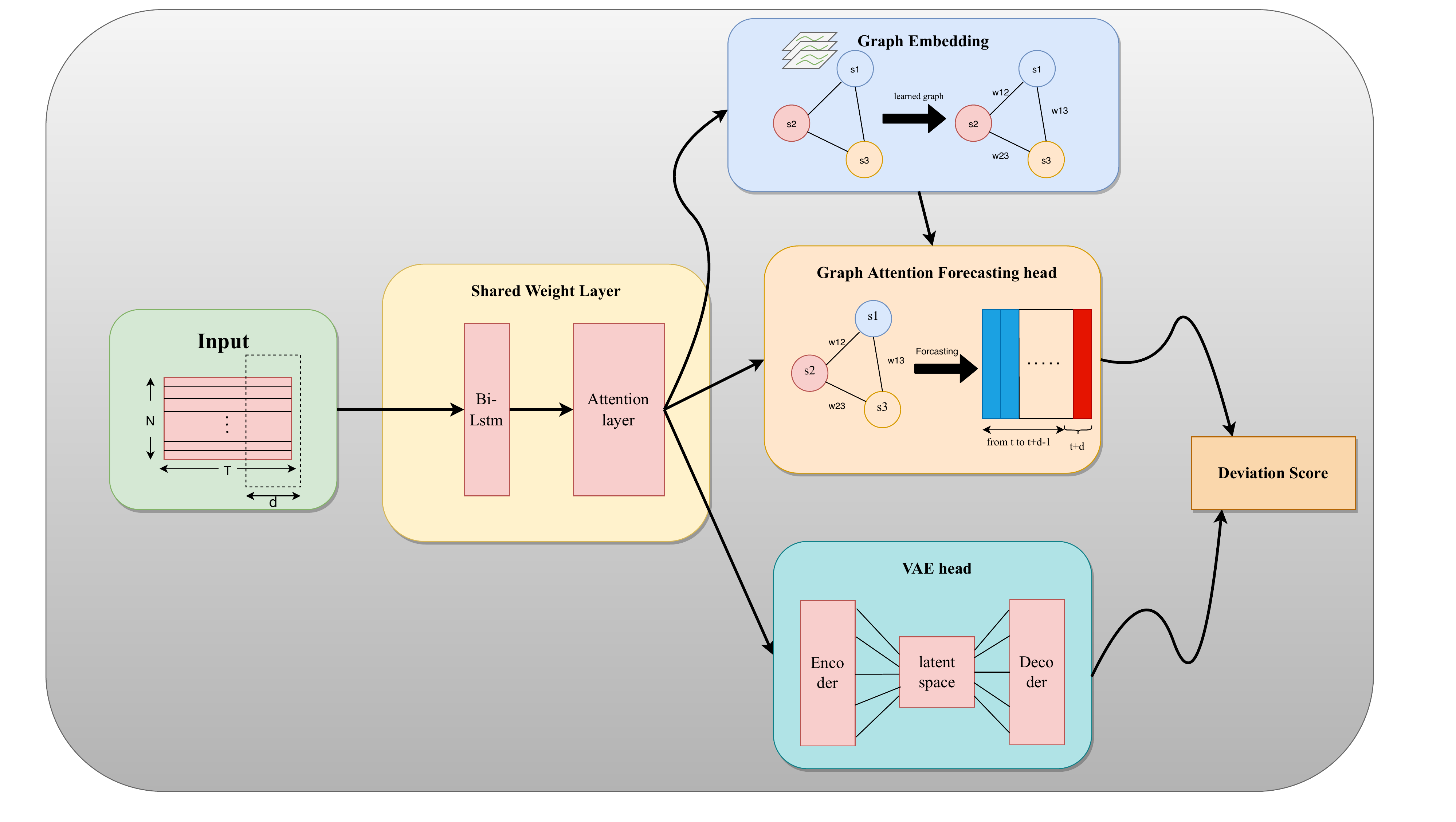}
\caption{Overview of MGADN.}\label{fig:MGADN}
\end{center}
\end{figure}

Our proposed model MGADN is composed of four major modules: \textbf{Shared Weight Layer}, \textbf{Graph Embedding Module}, \textbf{VAE head} and \textbf{Graph Attention Forecasting head}. The outputs of two heads will be transferred into deviation score and judged. Furthermore, because of our two-headed design, this model can be regraded as a multi-task learning framework. Due to this specified design, MGDA algorithm will be utilized in order to balance the training of two heads. Fig.~\ref{fig:MGADN} shows the overview framework of our proposed model.

\subsection{Shared Weighted Layer}
The first module of our model is  a layer whose weights are shared by the whole model. The shared layer is composed of a Bi-LSTM layer and a self attention layer. The framework of this layer is showed in fig2. Bi-LSTM acts as a time pattern collector, which can learn the relation within one slide window across time. When the number of hidden layer of Bi-LSTM is 1, the input $\left [  h^{t-d+1},h^{t-d+2},..., h^{t}\right ]$ will be transferred into:
\begin{equation}
\left [  \frac{(s_{forward}^{t-d+1}+s_{backward}^{t-d+1})}{2},  \frac{(s_{forward}^{t-d+2}+s_{backward}^{t-d+2})}{2}, ... , \frac{(s_{forward}^{t}+s_{backward}^{t})}{2}\right ]\nonumber,
\end{equation}

Attention technique is widely used in natural language processing task. For instance, in translation task, self attention can take full advantage of forward and backward context by setting attention weight. Attention weight describes the importance of input. The main step of attention technique is to find three variables which are Q(query), K(key), V(value). In our self attention layer, we assume $Q=K=V$.

Firstly, three weighting matrixes should be initialized  as $W^Q, W^K, W^V$. Then the input which is also the output of Bi-LSTM: $X$ will be operated below:
\begin{equation}
\left\{\begin{matrix}
 Q = W^Q X\\K = W^K X
 \\V = W^V X
\end{matrix}\right.
\end{equation}
\begin{equation}
Out = V \times Softmax(\frac{QK^T}{\sqrt{d_k}})
\end{equation}
\begin{figure}[htp]
\begin{center}
\includegraphics[width=1\textwidth]{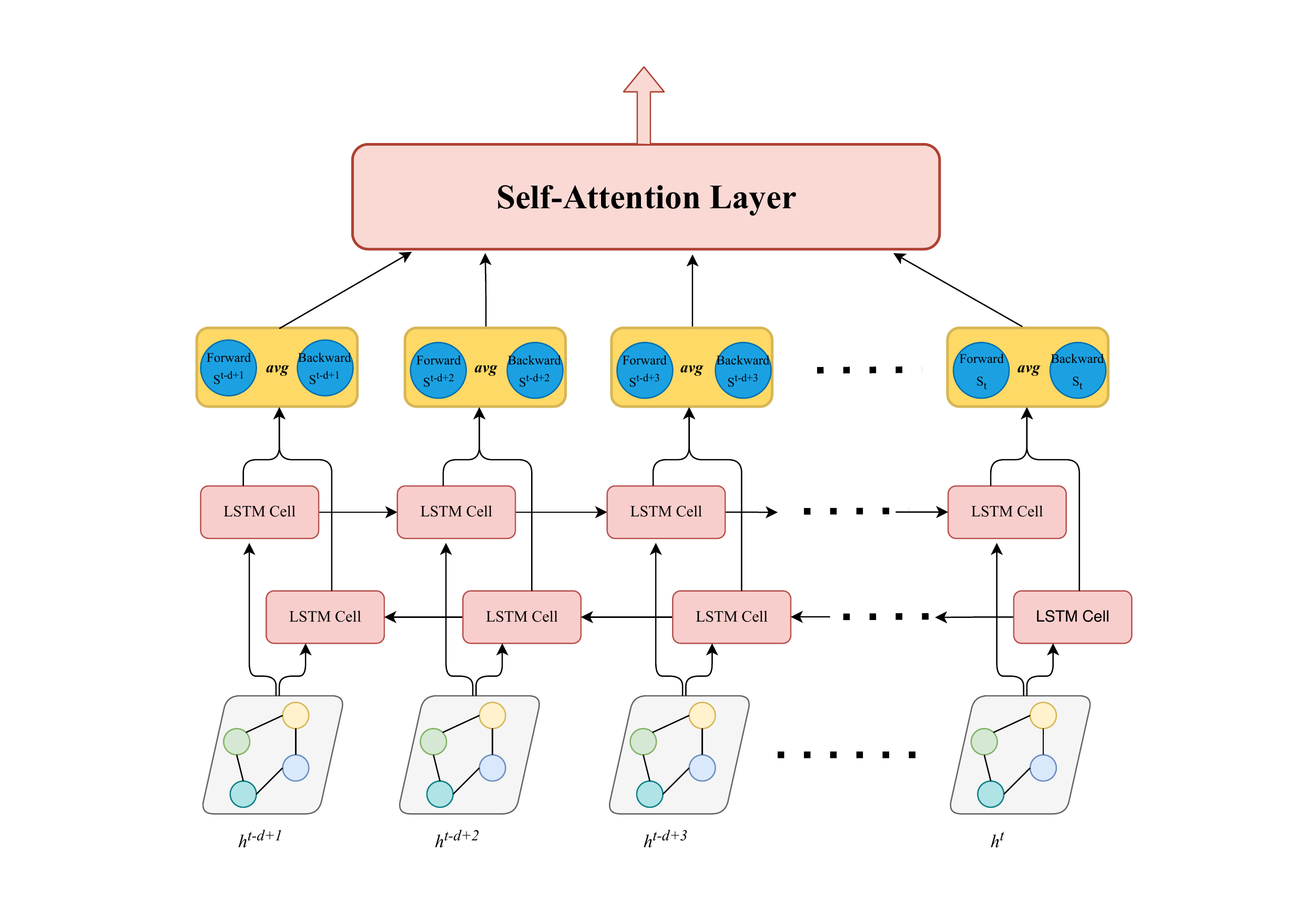}
\caption{Shared Weighted Layer, which is composed of two parts:Bi-Lstm and Self Attention Layer.}\label{fig:Shared}
\end{center}
\end{figure}
To clarify the design of shared weighted layer, Bi-LSTM layer capture forward and backward timely pattern. Self attention in turns reinforce some time stamps by giving them higher weight. In fact these two sub-modules play similar roles in this model. However, one is the other one's enhancement. The architecture of shared weighted layer is showed in Fig.~\ref{fig:Shared}.

\subsection{Graph Attention Forecasting Head}
\subsubsection{Graph Structure Learning}
As is mentioned above, we introduce GNNs to learn relationships between sensors, which brings another problem: how can we transfer multi-time series into graph structured data? \cite{wu2020connecting} in 2020 firstly propose a common solution to this problem by considering every sensors as nodes with its hidden states $S_i \in R^T ,i \in {1,2,3,...,N}$ i.e N sensors with T time stamps. Moreover, there are several ways of forming graph structure, which is, in other words, calculating adjacency matrix in different methods. For example, we can design directed graph by setting adjacency matrix asymmetric or undirected graph in similar way. Here, our solution is to acquire symmetric adjacency matrix by calculating Pearson correlation coefficient.

Provided with $N$ sensors, sensor $i$ is potentially related to the rest of all sensors. We define the potential neighbors of sensor $i$ as $P(i) = \left \{  1, 2,...,i-1,i+1,...,N\right\}$. The output of shared weighted layer will be initially sent to an embedding layer. The outputs of embedding layer are denoted as $v_i \text { for } i=1,2,3,...,N$. Calculations are as followed:
\begin{equation}
% e_{ji} = \frac{vi^T v^j}{\left \| v^i \right \| \left \| v^j \right \|} 
% for  j \in P(i)
e_{j i}=\frac{\mathbf{v}_{\mathbf{i}}^{\top} \mathbf{v}_{\mathbf{j}}}{\left\|\mathbf{v}_{\mathbf{i}}\right\| \cdot\left\|\mathbf{v}_{\mathbf{j}}\right\|} \text { for } j \in \mathcal{C}_{i}
\end{equation}
\begin{equation}
A_{j i}=1\left\{j \in \operatorname{TopK}\left(\left\{e_{k i}: k \in \mathcal{C}_{i}\right\}\right)\right\}    
\end{equation}
That is to say, we firstly calculate the Pearson correlation coefficient between every two sensors. Secondly, for a certain sensor, we leave the most related k sensors as its neighbors.

\subsubsection{Graph Attention Layer}
In this section, we bring in a specified Graph Attention Layer proposed by \cite{deng2021graph}.

The input of this module is $\mathbf{x}^{(t)}:= \left [  h^{t-d},h^{t-d+1},..., h^{t-1}\right ]$ when slide window is of $d$ length. And we need to give the prediction of $h^{t}$. We denote that $x^{(t)}_{i} \in R^d$ is input feature of node $i$, $\mathcal{N}(i)=\left\{j \mid A_{j i}>0\right\}$ is the neighborhood of node $i$ according to adjacency $A$. We have graph attention mechanism:
\begin{equation}
\begin{aligned}
\mathbf{g}_{i}^{(t)} &=\mathbf{v}_{i} \oplus \mathbf{W} \mathbf{x}_{i}^{(t)}
\end{aligned}
\end{equation}

\begin{equation}
\begin{aligned}
\pi(i, j) &=\operatorname{LeakyReLU}\left(\mathbf{a}^{\top}\left(\mathbf{g}_{i}^{(t)} \oplus \mathbf{g}_{j}^{(t)}\right)\right) 
\end{aligned}
\end{equation}
\begin{equation}
\begin{aligned}
\alpha_{i, j} &=\frac{\exp (\pi(i, j))}{\sum_{k \in \mathcal{N}(i) \cup\{i\}} \exp (\pi(i, k))},
\end{aligned}
\end{equation}
\begin{equation}
\begin{aligned}
\mathbf{z}_{i}^{(t)}=\operatorname{ReLU}\left(\alpha_{i, i} \mathbf{W} \mathbf{x}_{i}^{(t)}+\sum_{j \in \mathcal{N}(i)} \alpha_{i, j} \mathbf{W} \mathbf{x}_{j}^{(t)}\right)
\end{aligned}
\end{equation}
where $W \in R^{w \times d}$ is a trainable weight matrix which is also the embedding matrix before Graph Structure Learning layer. This weight matrix applies a learnable linear transformation towards every node. $\alpha_{i, j}$ is the attention weight between node $i$ and node $j$. Then, equation (8) provides iteration rule, which in turns let us obtain the representation of all nodes, namely $\left\{\mathbf{Z}_{1}^{(t)}, \cdots, \mathbf{Z}_{N}^{(t)}\right\}$. The output of Graph Attention Layer can be described as:
\begin{equation}
\begin{aligned}
\hat{\mathbf{h}}^{(\mathbf{t})}=f_{\theta}\left(\left[\mathbf{v}_{1} \circ \mathbf{z}_{1}^{(t)}, \cdots, \mathbf{v}_{N} \circ \mathbf{z}_{N}^{(t)}\right]\right)
\end{aligned}
\end{equation}
where $\circ$ means element-wise multiply, $f_{\theta}$ means stacked fully-connected layers with output dimension $N$.

\subsection{VAE Head}
VAE Head, which acts as a reconstruction model, is a Variable Auto Encoder module which is a updated version of AE(Auto Encoder). The underlying logic that AE can be competent for multi-time series anomaly detection is that compared with normal data, abnormal data is more difficult to be represented by low-dimensional latent layer, so when abnormal data is reconstructed, its reconstruction error will increase significantly, so as to determine the occurrence of abnormality. Furthermore, VAE encode input as a distribution while AE only encode input as a point in the latent space. This uncertain probability distribution provides more robust representation than which is provided by certain point. The whole process of VAE is as followed:
\begin{equation}
\begin{aligned}
\mu, \sigma=W_{\mu} g_{\theta_{1}}(X), W_{\sigma} g_{\theta_{1}}(X)
\end{aligned}
\end{equation}
\begin{equation}
\begin{aligned}
Z=\mu+\sigma \times \epsilon
\end{aligned}
\end{equation}
\begin{equation}
\begin{aligned}
\hat{X}=f_{\theta_{2}}(Z)
\end{aligned}
\end{equation}

\noindent Equation (10) is the process of encoder which transfer input data $X$ into a Gaussian Distribution with its own expectation and variance. Equation (11) is to find latent variable from this distribution. Equation (12) is decoder process. The encoder and decoder processes are both realized by trainable linear transformation.

\subsection{Multi-task Learning and Deviation Score}
\textbf{Multi-task Learning:}  At last, the outputs of two heads will be the output of prediction head: $\hat{\mathbf{h}}^{(\mathbf{t})}$, and the output of reconstruction head: $\left [  \hat{\mathbf{r}}^{(\mathbf{t-d})}, \hat{\mathbf{r}}^{(\mathbf{t-d+1})},..., \hat{\mathbf{r}}^{(\mathbf{t-1})}\right ] $ with respect to the input:$\left [  {\mathbf{h}}^{(\mathbf{t-d})}, {\mathbf{h}}^{(\mathbf{t-d+1})},..., {\mathbf{h}}^{(\mathbf{t-1})}\right ] $ and ground truth at time $t$: ${\mathbf{h}}^{(\mathbf{t})}$. The loss function is:
\begin{equation}
\begin{aligned}
L_{\mathrm{pred}}=\frac{1}{T_{\text {train }}-d} \sum_{t=d+1}^{T_{\text {train }}}\left\|\hat{\mathbf{h}}^{(\mathbf{t})}-\mathbf{h}^{(\mathbf{t})}\right\|_{2}^{2}
\end{aligned}
\end{equation}

\begin{equation}
\begin{aligned}
L_{\mathrm{recon}}=K L\left(N\left(\mu, \sigma^{2}\right) \| N(0, I)\right) + \frac{1}{T_{\text {train }}-d} \sum_{t=d+1}^{T_{\text {train }}}\left|\hat{\mathbf{r}}^{(\mathbf{t})}-\mathbf{h}^{(\mathbf{t})}\right|_{1}\\=
\frac{1}{2}\left(-\log \sigma^{2}+\mu^{2}+\sigma^{2}-1\right) + \frac{1}{T_{\text {train }}-d} \sum_{t=d+1}^{T_{\text {train }}}\left|\hat{\mathbf{r}}^{(\mathbf{t})}-\mathbf{h}^{(\mathbf{t})}\right|_{1}
\end{aligned}
\end{equation}

It should be pointed out that the original loss function of VAE will vibrate so much during training which causes the decrease of accuracy. Therefore, we replace Binary Cross Entropy with L1 loss, which in turns solve this problem. Obviously, the model has become a multi-task learning framework(MTL). To take full advantage of this MTL structure,
MGDA(Multiple Gradient Descent Algorithm) proposed by \cite{sener2018multi} can be used for reference.

When faced with multiple distinct loss functions, there is an intuitionistic way. That is, assign different weights to each loss function. Given $M$ different tasks, the overall loss function will be:
\begin{equation}
\begin{aligned}
\min _{\substack{\boldsymbol{\theta}^{s h}, \boldsymbol{\theta}^{1}, \ldots, \boldsymbol{\theta}^{M}}} \sum_{t=1}^{M} c^{m} \hat{\mathcal{L}}^{m}\left(\boldsymbol{\theta}^{s h}, \boldsymbol{\theta}^{m}\right)
\end{aligned}
\end{equation}
where $\theta^{s h}$ is parameters shared by all tasks while $\theta^{m}$ is the task specific parameters. Consider two sets of solutions $\theta$ and $\overline{\theta}$ such that $\hat{\mathcal{L}}^{m_{1}}\left(\boldsymbol{\theta}^{s h}, \boldsymbol{\theta}^{m_{1}}\right)<\hat{\mathcal{L}}^{m_{1}}\left(\overline{\boldsymbol{\theta}}^{s h}, \overline{\boldsymbol{\theta}}^{m_{1}}\right)$ and $\hat{\mathcal{L}}^{m_{2}}\left(\boldsymbol{\theta}^{s h}, \boldsymbol{\theta}^{m_{2}}\right)>\hat{\mathcal{L}}^{m_{2}}\left(\overline{\boldsymbol{\theta}}^{s h}, \overline{\boldsymbol{\theta}}^{m_{2}}\right)$  for some tasks $m_1$ and $m_2$. In other words, solution $\theta$ is better for task $m_1$ whereas $\overline{\theta}$ is better for $m_2$. It is not possible to compare these two solutions without a pairwise importance of tasks, which is typically not available. To solve this problem, MGDA provides us with a alternative solution by calculating Pareto solution. Firstly, we can give the definition of Pareto optimality for MTL.

The optimization goal becomes:
\begin{equation}
\min _{\substack{\boldsymbol{\theta}^{s h}, \boldsymbol{\theta}^{1}, \ldots, \boldsymbol{\theta}^{M}}} \mathbf{L}\left(\boldsymbol{\theta}^{s h}, \boldsymbol{\theta}^{1}, \ldots, \boldsymbol{\theta}^{T}\right)=\min _{\substack{\boldsymbol{\theta}^{s h}, \boldsymbol{\theta}^{1}, \ldots, \boldsymbol{\theta}^{M}}}\left(\hat{\mathcal{L}}^{1}\left(\boldsymbol{\theta}^{s h}, \boldsymbol{\theta}^{1}\right), \ldots, \hat{\mathcal{L}}^{M}\left(\boldsymbol{\theta}^{s h}, \boldsymbol{\theta}^{M}\right)\right)^{\top}    
\end{equation}

\noindent\textbf{Definition 2: Pareto optimality for MTL}
\begin{itemize}
\item A solution $\theta$ dominates $\overline{\theta}$ if $\hat{\mathcal{L}}^{m}\left(\boldsymbol{\theta}^{s h}, \boldsymbol{\theta}^{m}\right) \leq \hat{\mathcal{L}}^{m}\left(\overline{\boldsymbol{\theta}}^{s h}, \overline{\boldsymbol{\theta}}^{m}\right)$ for all task $m$ and satisfying $\mathbf{L}\left(\boldsymbol{\theta}^{s h}, \boldsymbol{\theta}^{1}, \ldots, \boldsymbol{\theta}^{M}\right) \neq \mathbf{L}\left(\overline{\boldsymbol{\theta}}^{s h}, \overline{\boldsymbol{\theta}}^{1}, \ldots, \overline{\boldsymbol{\theta}}^{M}\right)$.

\item A solution $\tilde{\theta}$ is called Pareto optimal if there is no other solutions dominate $\tilde{\theta}$.
\end{itemize}

\noindent Considering the Karush-Kuhn-Tucker conditions for both shared and task specific parameters, the optimization task becomes:
\begin{equation}
    \min _{\alpha^{1}, \ldots, \alpha^{M}}\left\{\left\|\sum_{m=1}^{M} \alpha^{m} \nabla_{\boldsymbol{\theta}^{s h}} \hat{\mathcal{L}}^{m}\left(\boldsymbol{\theta}^{s h}, \boldsymbol{\theta}^{m}\right)\right\|_{2}^{2} \mid \sum_{m=1}^{M} \alpha^{m}=1, \alpha^{m} \geq 0 \quad \forall m\right\}
\end{equation}
\cite{desideri2012multiple} proved that either the solution to this optimization problem is 0 and the resulting point satisfies the KKT conditions, or the solution gives a descent direction that improves all tasks. Hence, the optimization problem becomes solving equation (17). \cite{sener2018multi} also come up with an approximation verson of MGDA, which is MGDA-UB by clarifying the inequation below:
\begin{equation}
    \left\|\sum_{m=1}^{M} \alpha^{m} \nabla_{\boldsymbol{\theta}^{s h}} \hat{\mathcal{L}}^{m}\left(\boldsymbol{\theta}^{s h}, \boldsymbol{\theta}^{m}\right)\right\|_{2}^{2} \leq\left\|\frac{\partial \mathbf{Z}}{\partial \boldsymbol{\theta}^{s h}}\right\|_{2}^{2}\left\|\sum_{m=1}^{M} \alpha^{m} \nabla_{\mathbf{Z}} \hat{\mathcal{L}}^{m}\left(\boldsymbol{\theta}^{s h}, \boldsymbol{\theta}^{m}\right)\right\|_{2}^{2}
\end{equation}
where $Z$ is the output of shared weighted layer. Then, the aproximation of MGDA becomes:
\begin{equation}
    \min _{\alpha^{1}, \ldots, \alpha^{M}}\left\{\left\|\sum_{m=1}^{M} \alpha^{m} \nabla_{\mathbf{Z}} \hat{\mathcal{L}}^{m}\left(\boldsymbol{\theta}^{s h}, \boldsymbol{\theta}^{m}\right)\right\|_{2}^{2} \mid \sum_{m=1}^{M} \alpha^{m}=1, \alpha^{m} \geq 0 \quad \forall m\right\}
\end{equation}

\noindent When it comes to our model, the optimization problem is:
\begin{equation}
    \min _{\alpha \in[0,1]}\left\|\alpha \nabla_{\boldsymbol{\theta}^{s h}} \hat{\mathcal{L}}_{pred}\left(\boldsymbol{\theta}^{s h}, \boldsymbol{\theta}^{1}\right)+(1-\alpha) \nabla_{\boldsymbol{\theta}^{s h}} \hat{\mathcal{L}}_{recon}\left(\boldsymbol{\theta}^{s h}, \boldsymbol{\theta}^{2}\right)\right\|_{2}^{2}
\end{equation}
Then, according to MGDA-UB, equation (20) becomes:
\begin{equation}
    \min _{\alpha \in[0,1]}\left\|\alpha \nabla_{\boldsymbol{Z}} \hat{\mathcal{L}}_{pred}\left(\boldsymbol{\theta}^{s h}, \boldsymbol{\theta}^{1}\right)+(1-\alpha) \nabla_{\boldsymbol{Z}} \hat{\mathcal{L}}_{recon}\left(\boldsymbol{\theta}^{s h}, \boldsymbol{\theta}^{2}\right)\right\|_{2}^{2}
\end{equation}

\noindent Equation (21) is a dimensional quadratic function of $\alpha$ which has an analytical solution:
\begin{equation}
    a =\left[\frac{\left(\nabla_{\boldsymbol{Z}} \hat{\mathcal{L}}_{recon}\left(\boldsymbol{\theta}^{sh}, \boldsymbol{\theta}_{recon}\right)-\nabla_{\boldsymbol{Z}} \hat{\mathcal{L}}_{pred}\left(\boldsymbol{\theta}^{sh}, \boldsymbol{\theta}_{pred}\right)\right)^{\top} \nabla_{\boldsymbol{Z}} \hat{\mathcal{L}}_{recon}\left(\boldsymbol{\theta}^{sh}, \boldsymbol{\theta}_{recon}\right)}{\left\|\nabla_{\boldsymbol{Z}} \hat{\mathcal{L}}_{pred}\left(\boldsymbol{\theta}^{sh}, \boldsymbol{\theta}_{pred}\right)-\nabla_{\boldsymbol{Z}} \hat{\mathcal{L}}_{recon}\left(\boldsymbol{\theta}^{sh}, \boldsymbol{\theta}_{recon}\right)\right\|_{2}^{2}}\right]
\end{equation}
\begin{equation}
    \hat{\alpha} = max(min(a, 1),0)
\end{equation}

\begin{equation}
    Loss = \hat{\alpha} \mathcal{L}_{pred} + (1-\hat{\alpha}) \mathcal{L}_{recon}
\end{equation}

\noindent\textbf{Deviation Score:} After acquiring the output of two heads, the deviation score calculation will be:
\begin{equation}
    \operatorname{Err}_{i}^{pred}(t)=\left|\mathbf{h}_{\mathbf{i}}^{(\mathbf{t})}-\hat{\mathbf{h}}_{\mathbf{i}}^{(\mathbf{t})}\right|
\end{equation}
\begin{equation}
    \operatorname{Err}_{i}^{recon}(t)=\left|\mathbf{h}_{\mathbf{i}}^{(\mathbf{t})}-\hat{\mathbf{r}}_{\mathbf{i}}^{(\mathbf{t})}\right|
\end{equation}
\begin{equation}
    a_{i}^{pred}(t)=\frac{\operatorname{Err}_{i}^{pred}(t)-\widetilde{\mu}_{i}^{pred}}{\widetilde{\sigma}_{i}^{pred}}
\end{equation}
\begin{equation}
    a_{i}^{recon}(t)=\frac{\operatorname{Err}_{i}^{recon}(t)-\widetilde{\mu}_{i}^{recon}}{\widetilde{\sigma}_{i}^{recon}}
\end{equation}
\begin{equation}
    A(t)=\max (\max _{i} a_{i}^{pred}(t), \max _{i} a_{i}^{recon}(t))
\end{equation}

\noindent where $\widetilde{\mu}_{i},\widetilde{\sigma}_{i}$ are the median and inter-quartile range across time stamps with respect to $\operatorname{Err}_{i}(t)$. $A(t)$ is calculated as the maximum between prediction score and reconstruction score, by which model can retain its robustness and becomes more sensitive towards anomalies.

At last, a time stamp $t$ is labelled as anomaly when $A(t)$ exceed a certain threshold, which is set to be the max of $A(t)$ within the validation data.

\section{Experiment}
In this section, we will proceed several experiments for MGADN on three open datasets. Moreover, in order to clarify interpretability of our model, ablation study and some visualization works will be done.
\subsection{Datasets}
We first introduce datasets on which we conduct our experiments:
\begin{itemize}
    \item \textbf{MSL:} This data represents real spacecraft telemetry data and anomalies from the Curiosity Rover on Mars (MSL). 
    \item \textbf{SWaT:} The Secure Water Treatment (SWaT) dataset is from a water treatment test platform coordinated by Singapore’s Public Utility Board \cite{mathur2016swat}.
    \item \textbf{WADI:}  Water Distribution (WADI) \cite{ahmed2017wadi} is a distribution system comprising a larger number of water distribution pipelines which is the extension of SWaT.
\end{itemize}

\begin{table*}[t]\footnotesize
\setlength{\abovecaptionskip}{0.1cm}
\caption{The Properties of Datasets}
%\vspace{-0.1in}
%\vspace{-0.05in} 
\centering
%\resizebox{\textwidth}{!}{
\begin{tabular}{l|rrrr}
\hline  %添加表格头部粗线                
Datasets   & Features    & Train  & Test  & Anomalies         \\ \hline  %添加表格中横线
MSL   &27&1564&2048&21.90$\%$\\
SWaT  &51&49678&44991&12.13$\%$\\
WADI  &127&78457&17280&5.83$\%$\\
\hline
\end{tabular}
%}
\vspace{-1.em}
\label{CIFAR100-LT}
\end{table*}

\subsection{Baselines}
\begin{itemize}
    \item \textbf{PCA (Principal Component Analysis):} finds a low-dimensional projection that keeps most of the variance within the dataset. The deviation score is the reconstruction error of this projection \cite{shyu2003novel}.
    \item \textbf{KNN (K Nearest Neighbors):} uses each point’s distance to its kth nearest neighbor as an deviation score \cite{angiulli2002fast}.
    \item \textbf{AE (Autoencoders):} An encoder-decoder model which uses reconstruction error as its deviation score \cite{aggarwaloutlier}.
    \item \textbf{DAGMM (Deep Autoencoding Gaussian Model):} combines  Gaussian Mixture Model with AE. It uses reconstruction error as its deviation score \cite{zhou2019beatgan}.
    \item \textbf{LSTM-VAE:} replaces the encoder and decoder which is composed of linear layers in VAE with LSTM. It uses reconstruction error as its deviation score \cite{park2018multimodal}.
    \item \textbf{MAD-GAN:} A GAN-like model, which uses LSTM-RNN as discriminator along with a reconstruction based anomaly detection approach \cite{li2019mad}.
    \item \textbf{GDN:} A prediction model, which combines GAT with anomaly detection approach and calculate deviation score at every predicted time stamps \cite{deng2021graph}.
    
\end{itemize}
\subsection{Setup}
We program our model with Pytorch version 1.5.1 with Cuda 11.1. PyTorch Geometric Library version 1.5.0 \cite{fey2019fast}. The model is trained on Intel(R) Xeon(R) CPU E5-2699 v3 @ 2.30GHz and RTX2080TI. The models are trained using the Adam optimizer with learning rate 0.001. The number of epochs are at most 50. And we choose the number of neighbors k as 5(15, 30) for MSL(SWaT, WADI). The length of slide window will be 5 for all datasets.

\subsection{Metrics:}
In this section, we will introduce the evaluation metrics that is used in experiments. We use precision(prec), recall(rec) and F1 score(F1) on the test data and its ground truth to measure the capability of models. $Prec = \frac{TP}{TP+FP}$, $Rec = \frac{TP}{TP+FN}$ and $F1 = \frac{2Prec \times Rec}{Prec + Rec}$, where TN, TP, FN and FP represent true negative, true positive, false negative and false positive.

\begin{table*}[t]\footnotesize
\setlength{\abovecaptionskip}{0.1cm}
\caption{Experiment Result of Precision($\%$), Recall($\%$) and F1-score.}
%\vspace{-0.05in} 
%\vspace{-0.1in}
\centering
\resizebox{\textwidth}{!}{
\begin{tabular}{l|rrr|rrr|rrr}
\hline  %添加表格头部粗线
Dataset &\multicolumn{3}{c|}{MSL} &\multicolumn{3}{c}{SWaT} &\multicolumn{3}{c}{WADI}\\ \hline
                 & Prec.   & Rec.   & F1.   & Prec.   & Rec.   & F1.  & Prec.   & Rec.   & F1.     \\ \hline  %添加表格中横线
PCA &24.53&16.34&0.10  &24.92 &21.63 &0.23 &39.53 &5.63 &0.10\\
KNN &10.96&10.96&0.08  &7.83 &7.83 &0.08 &7.76 &7.75 &0.08\\
AE  &53.92 &67.63 &0.60 &72.63 &52.63 &0.61 &34.35 &34.35 &0.34\\
DAGMM  &49.24 &70.32 &0.29 &27.46 &69.52 &0.39 &54.44 &26.99 &0.36\\
LSTM-VAE &69.82 &79.93 &0.37 &96.24 &59.91 &0.74 &87.79 &14.45 &0.25\\
MAD-GAN &70.67 &82.33 &0.76  &98.97 &63.74 &0.77 &41.44 &33.92 &0.37\\
GDN &79.08 &\textbf{99.50} &0.88  &99.35 &68.12 &0.81 &\textbf{97.50} &40.19 &0.57\\
\hline
\textbf{MGADN-ALT} &78.31&98.81&0.87 &98.45&69.44&0.84 &92.97&49.33&0.64\\
\textbf{MGADN-MGDA} &\textbf{80.80}&98.87&\textbf{0.89} &\textbf{99.50}&\textbf{76.95}&\textbf{0.87} &94.16&\textbf{50.98}&\textbf{0.66}\\

\hline %添加表格底部粗线
\end{tabular}}
\vspace{-1.em}
\label{results}
\end{table*}

\subsection{Experiment Results}
\subsubsection{Accuracy}
In table.~\ref{results}, the results of experiments show that MGADN outperforms baselines in terms of precision, recall and F1-score. We also use two different ways to train MGADN: with MGDA and without MGDA (i.e MGADN-MGDA and MGADN-ALT). To specify, we train two heads alternately in MGADN-ALT. The result shows that MGDA-UB algorithm do make contributions in the model.
\subsubsection{Ablation Study}
In this section, we proceed ablation study on dataset SWaT to justify the necessity of each module. The result shows that without MGDA-UB to balance two losses from two heads, the precision will slightly decrease but recall will decrease more. Dropping VAE head will cause decrease of recall. Deleting predicted head will cause decrease of all three metrics. Table.~\ref{SWaT-ablation} shows the results of ablation study.
\begin{table*}[t]\footnotesize
\setlength{\abovecaptionskip}{0.1cm}
\caption{Abliation study on SWaT.}
\centering
\begin{tabular}{l|rrr}
\hline  %添加表格头部粗线
Model                 & Prec.    & Rec.    & F1.        \\ \hline  %添加表格中横线

MGADN &\textbf{99.50}&\textbf{76.95}&\textbf{0.87}\\
MGADN w/o MGDA-UB(i.e MGADN-ALT) &98.45&69.44&0.84\\
MGADN w/o VAE head &98.97&72.56&0.81\\
MGADN w/o predicted head &77.89&58.99&0.67\\
MGADN w/o VAE head w/o Shared weighted layer(i.e GDN) &99.35 &68.12 &0.81\\
MGADN w/o predicted head w/o Shared weighted layer(i.e VAE) &74.75 &57.79 &0.65\\
\hline
\end{tabular}
\vspace{-1.em}
\label{SWaT-ablation}
\end{table*}

\subsection{Other Interpretability}
In this section, some visualization work on MSL will be completed to better interpret the reliability of MGADN.

\begin{figure}
\begin{center}
\includegraphics[width=0.3\textwidth]{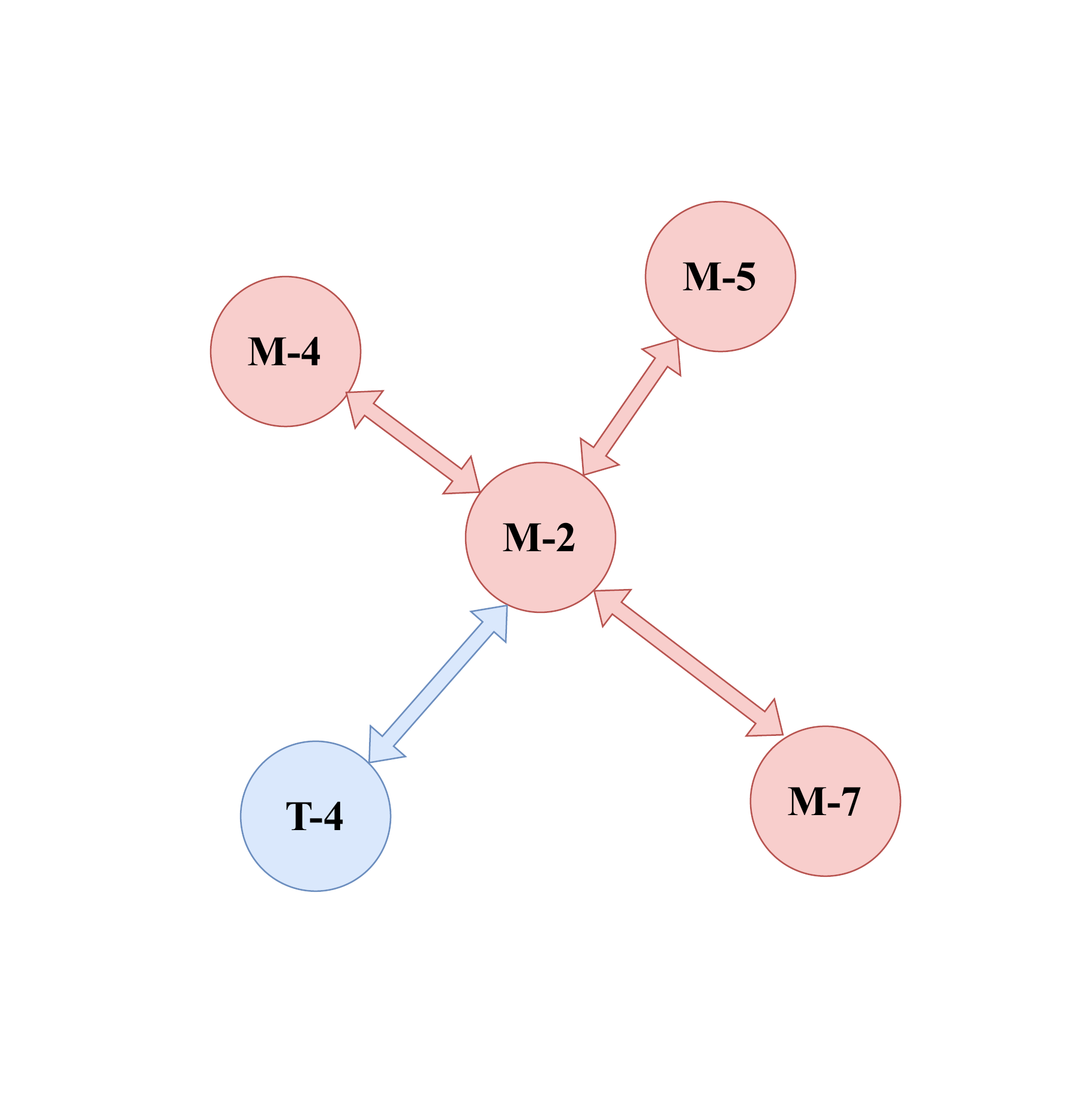}
\caption{Sensor M-2's neighbor learned by MGADN.Our model succeeded in capturing the relationship between M-7, M-4, M-5, which are highly correlated in reality. The potential relation between M-2 and T-4 is also revealed.}\label{fig:node}
\end{center}
\end{figure}

When we chose sensor M-2 as an example, according to the topk-masked adjacent matrix, M-4, M-7, M-5 are regarded as the neighborhood of M-2. The description of the dataset tells us that there are actually connections between these sensors because they are sensors that receive the similar signals on Curiosity Rover. Potential connection between T-4 and M-2 are also revealed. Fig.~\ref{fig:node} shows this situation.

\begin{figure}[]
\centering
\includegraphics[width=1\textwidth]{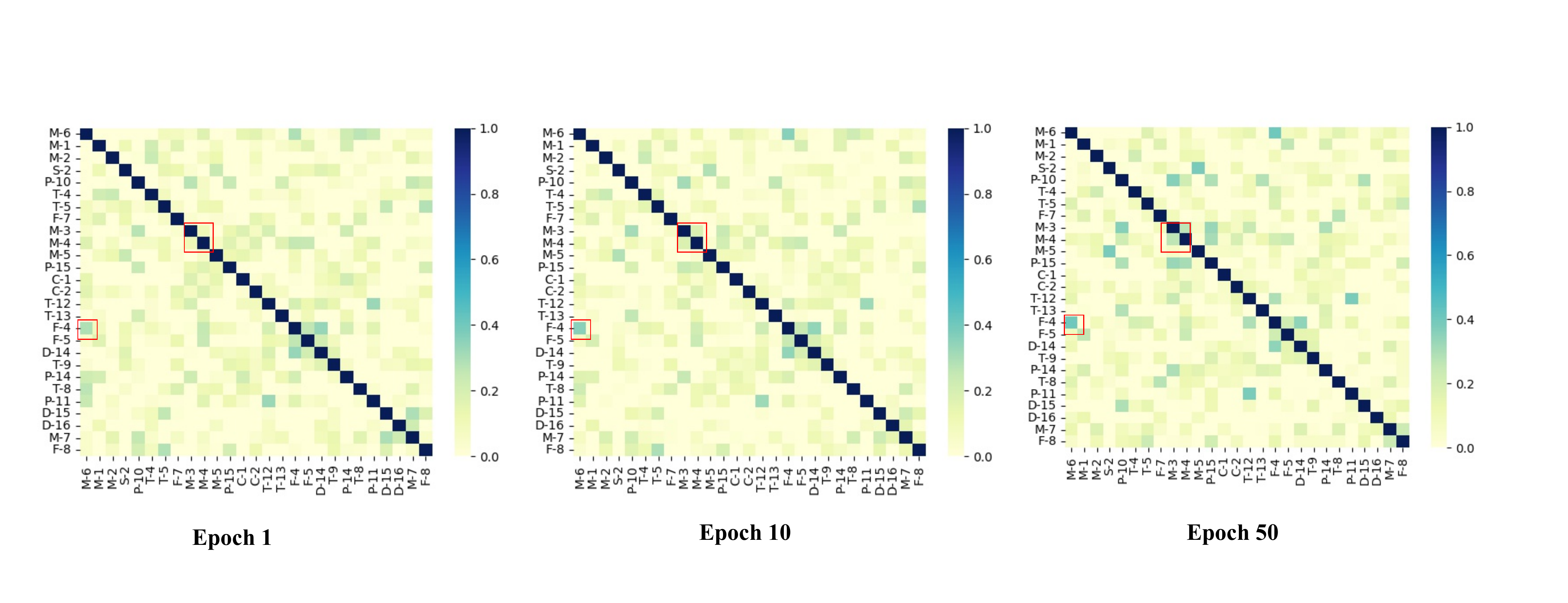}
\caption{The adjacent matrix learned by MGADN without topk mask. As the number of epoch grows from 1 to 50, MGADN gradually capture the relationship between M-3 and M-4. Moreover, the relationship between sensors which are not intuitively connected is also learned such as F-4 and M-6. }\label{fig:adj}
\end{figure}

Fig.~\ref{fig:adj} shows the variation of the unmasked adjacent matrix during the growth of epoch.

\section{Conclusion}
In this paper, we propose Multi-task Graph Anomaly Detection Network(MGADN), which is a two-headed time series anomaly detection approach. MGADN is able to capture relation between sensors automatically without any prior information. Multi-task optimization theory is also considered when tackling with losses of two different heads. Experiments on three real-world datasets proves its competence. 

Future work may consider different kinds of reconstruction model or better graph embedding approach when learning graph structure. For instance, reconstruction models like GAN or LSTM-VAE may be introduced to multi-head framework. Instead of learning one-directed graph structure, we may also make it bi-directional via different graph embedding approach.

% An example of citation~\cite{DBLP:conf/acml/2009}.

% \acks{Acknowledgements should go at the end, before appendices and references. You can uncomment this for the camera-ready version on paper acceptance.}

%\bibliographystyle{plain}
\bibliography{acml22}

% \appendix\begin{center}
% \begin{equation}
% …

% \end{equation}

% \end{center}

% \section{First Appendix}\label{apd:first}

% This is the first appendix.

% \section{Second Appendix}\label{apd:second}

% This is the second appendix.

\end{document}